\documentclass[runningheads]{llncs}

\usepackage[final,year=2024,ID=*****]{eccv}

\usepackage{eccvabbrv}

\usepackage{graphicx}
\usepackage{booktabs}

\usepackage[accsupp]{axessibility}  

\usepackage[pagebackref,breaklinks,colorlinks,citecolor=eccvblue]{hyperref}

\usepackage{orcidlink}

\begin{document}

\title{Open-Vocabulary Object Detectors: Robustness Challenges under Distribution Shifts}

\titlerunning{Robustness Challenges of OV-OD}

\author{
Prakash Chandra Chhipa\inst{1,*}\orcidID{0000-0002-6903-7552} \and
Kanjar De\inst{2}\orcidID{0000-0003-0221-8268} \and
Meenakshi Subhash Chippa\inst{1,+}\orcidID{0009-0000-2770-6271} \and
Rajkumar Saini\inst{1}\orcidID{0000-0001-8532-0895} \and
Marcus Liwicki\inst{1}\orcidID{0000-0003-4029-6574}
}

\authorrunning{Chhipa et al.}

\institute{Luleå Tekniska Universitet, Sweden \\
\email{\{first.middle.last\}@ltu.se} \\
\email{+ meechi-2@student.ltu.se} \and
 Fraunhofer Heinrich-Hertz-Institut, Berlin, Germany \\
\email{kanjar.de@hhi.fraunhofer.de}\\
\email{*corresponding author - prakash.chandra.chhipa@ltu.se}
}

\maketitle

\begin{abstract}
  The challenge of Out-Of-Distribution (OOD) robustness remains a critical hurdle towards deploying deep vision models. Vision-Language Models (VLMs) have recently achieved groundbreaking results. VLM-based open-vocabulary object detection extends the capabilities of traditional object detection frameworks, enabling the recognition and classification of objects beyond predefined categories. Investigating OOD robustness in recent open-vocabulary object detection is essential to increase the trustworthiness of these models. This study presents a comprehensive robustness evaluation of the zero-shot capabilities of three recent open-vocabulary (OV) foundation object detection models: OWL-ViT, YOLO World, and Grounding DINO. Experiments carried out on the robustness benchmarks COCO-O, COCO-DC, and COCO-C  encompassing distribution shifts due to information loss, corruption, adversarial attacks, and geometrical deformation, highlighting the challenges of the model's robustness to foster the research in this field. Project webpage: \url{https://prakashchhipa.github.io/projects/ovod_robustness}
\end{abstract}

\section{Introduction}
\label{sec:intro}

Recent studies \cite{chhipa2023can} on self-supervised learning have highlighted significant performance impacts under distribution shifts and corruptions, urging enhanced robustness strategies. Similarly, the robustness of foundation models must be examined to understand their resilience against various distribution shifts, corruptions, and adversarial attacks. Foundation AI models, pre-trained on extensive datasets spanning multiple domains, are designed with the primary objective of acquiring a comprehensive understanding of the world and applying their knowledge effectively across a wide range of downstream tasks and applications, thereby facilitating advancements in AI capabilities across multiple fields.
 \begin{figure}[h]
  \centering
  \includegraphics[width=0.6\columnwidth]{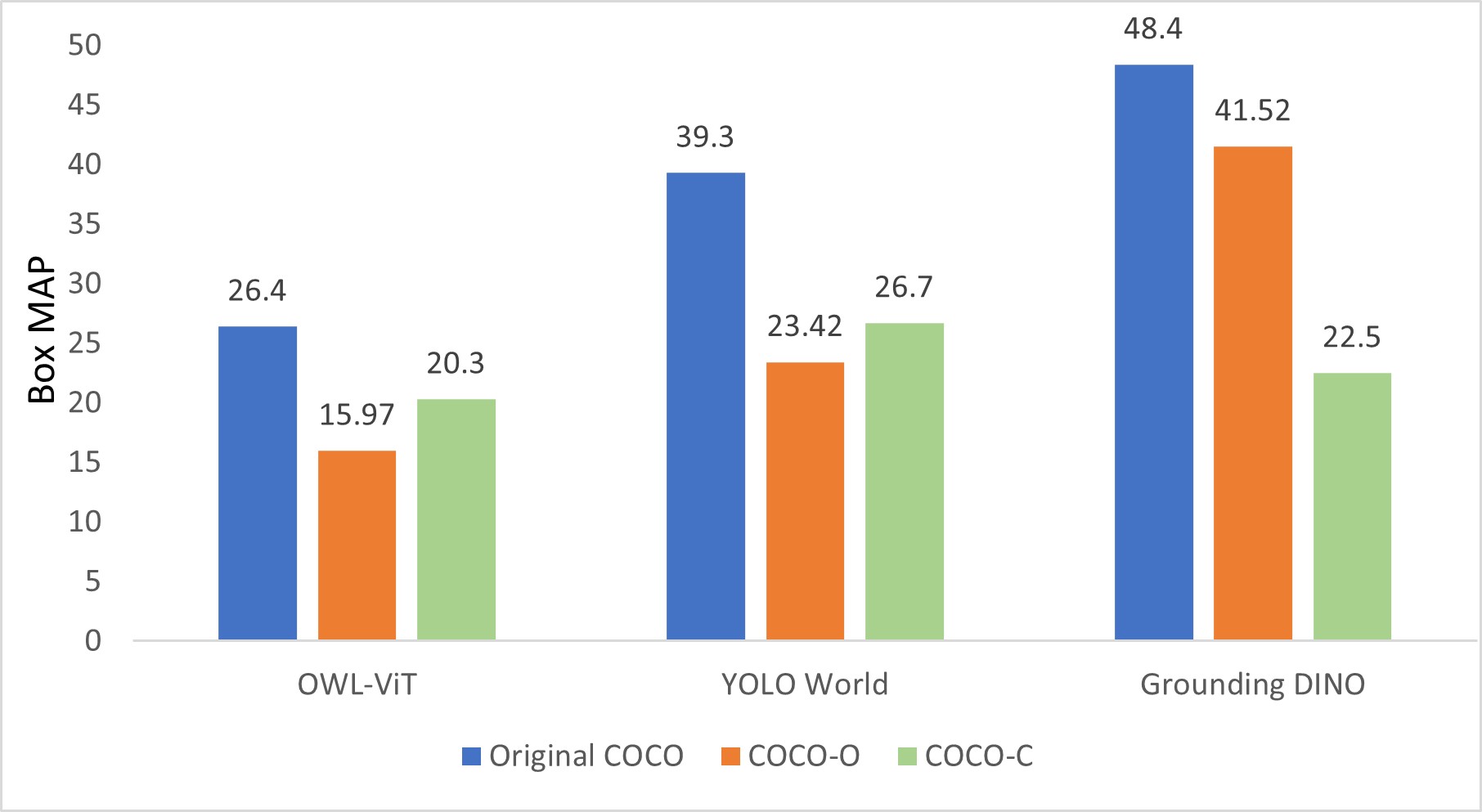}
  \caption{Zero-shot performance comparison for open vocabulary object detection models, OWL-ViT \cite{minderer2022simple}(ECCV'22), YOLO World \cite{cheng2024yolow} (CVPR'24), and Grounding DINO \cite{liu2023grounding} (ECCV'24). COCO-O \cite{mao2023coco} (ICCV'23) represents average results on six subsets, and COCO-C \cite{michaelis2019benchmarking} represents average results on fifteen corruptions.}
  \label{fig:meta}
\end{figure}

 \begin{figure}[h]
  \centering
  \includegraphics[width=0.8\columnwidth]{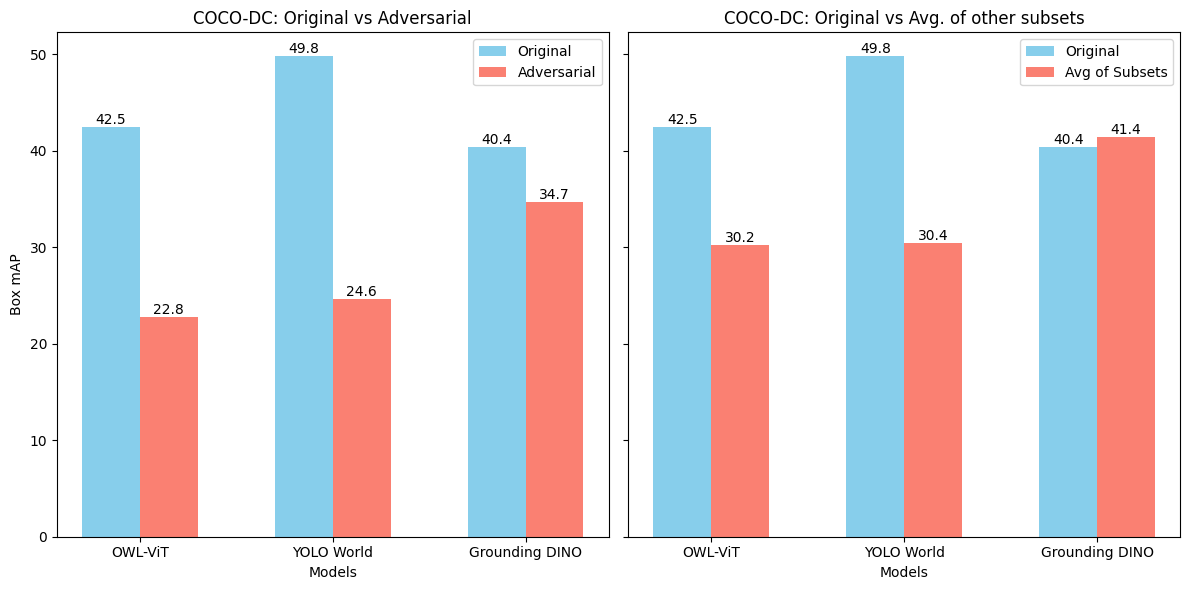}
  \caption{Zero-shot performance on COCO-DC: (left): comparison for OWL-ViT, YOLO World, and Grounding DINO on COCO-DC robustness performance on original subset and adversarial subset. (right):  comparison for these foundation models on COCO-DC robustness performance on original subset and average of all remaining subsets.}
  \label{fig:coco_dc_fig}
\end{figure}
The concept of foundation models has been most prominently developed in the fields of natural language processing (NLP) and computer vision (CV). In NLP, foundation models like Generative Pre-trained Transformer (GPT) \cite{achiam2023gpt} and Bidirectional Encoder Representations from Transformers (BERT) \cite{devlin2018bert} have revolutionized the field by enabling a range of applications, including text generation, sentiment analysis, question answering, and language translation, without the need for task-specific model architectures. 

Models such as Contrastive Language–Image Pre-training (CLIP)~\cite{radford2021learning} and DALL-E \cite{ramesh2021zero} demonstrate the ability to understand and generate visual content in response to natural language prompts, showcasing the versatility and creative potential of foundation models. Segment Anything Model (SAM) \cite{kirillov2023segment} is a foundation model for image segmentation. The utility of foundation models lies in their ability to leverage their pre-trained knowledge to perform a wide variety of tasks with minimal additional training. This versatility makes them a powerful tool for developing AI applications quickly and efficiently, opening up new possibilities for innovation and research in AI. 
 \begin{figure*}[!t]
  \centering
  \includegraphics[width=.8\columnwidth]{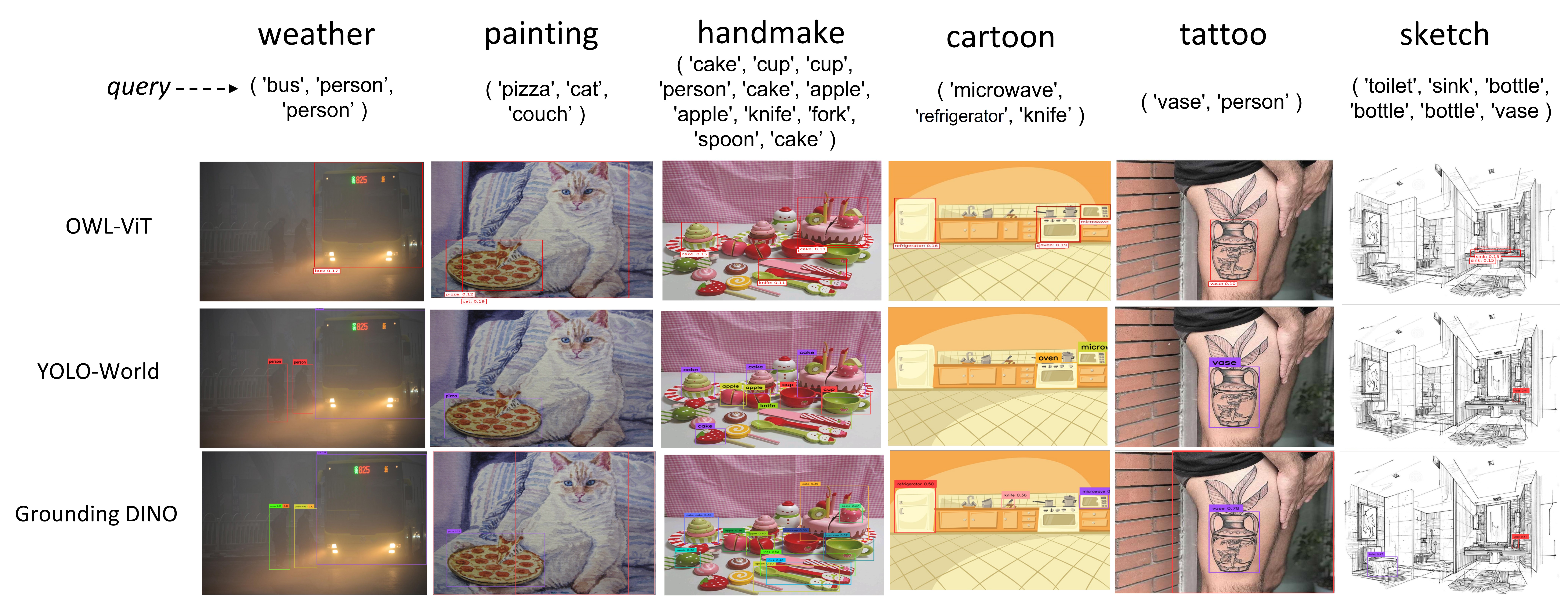}
  \caption{Examples from six COCO-O benchmark subsets depicted with predictions by open-vocabulary models: OWL-ViT, YOLO World, and Grounding DINO. The input textual query includes the object categories identified in the labels.}
  \label{fig:cartoon}
\end{figure*}
Open-vocabulary object detection has gained researchers' interest in solving real-world problems. Recent advances in open-vocabulary (OV) object detection in computer vision extend the capabilities of traditional object detection frameworks to recognize and classify objects beyond predefined categories present in their training datasets. Open-vocabulary (OV) models demonstrated zero-shot or few-shot learning capabilities by using vision-language pre-training, making the models to accurately identify and categorize objects they have never encountered during training. 

Robustness, defined as a model's ability to maintain consistent performance under varying and unforeseen conditions, has emerged as a critical factor in evaluating the utility of modern machine learning models across diverse applications and data distributions. These conditions include but are not limited to, noisy data, distribution shifts, and adversarial attacks. Robustness involves balancing the trade-off between model accuracy and the ability to generalize well across unforeseen scenarios, challenging the traditional focus on maximizing dataset-specific performance. Robust models contribute to enhanced security, privacy, and user trust, necessitating strategies encompassing data augmentation, regularization techniques, ensemble learning, and more to address challenges like adversarial robustness and long-tail robustness \cite{hendrycks2021unsolved}. 

We explore the robustness of foundation models, with a particular focus on open-vocabulary models. The connection between robustness and the trustworthiness of these models is crucial; robust models inspire greater confidence in their application across various domains, especially when leveraging zero-shot and few-shot learning capabilities. By bringing attention to the topic of robustness of open-vocabulary models, we inspire to improve trustworthiness by investigating the out-of-distribution performance on zero-shot evaluation.

To the best of our knowledge, we are the first study to provide comparative analysis of recent state-of-the-art OV object detectors: OWL-ViT \cite{minderer2022simple}, YOLO-World \cite{cheng2024yolow}, and Grounding DINO \cite{liu2023grounding} for open vocabulary object detection through the lenses of robustness. OWL-ViT leverages the Vision Transformer (ViT) for transfer learning, YOLO-World builds on the efficient and practical CNN architecture, and Grounding DINO integrates the Swin Transformer \cite{liu2021swin} with a novel grounded pre-training strategy.

We employ a zero-shot evaluation approach (Refer Figure \ref{fig:cartoon}), where models are tested on the out-of-distribution benchmarks such as COCO-O \cite{mao2023coco} and COCO-C \cite{michaelis2019benchmarking}, without any additional training on its specific subsets. COCO-O offers six subsets with decreasing details of objects in terms of shape, color, and textures, whereas COCO-C comprises fifteen subsets corresponding to corruptions from \cite{hendrycks2018benchmarking}. This evaluation investigates the models' ability to generalize from learned representations to unseen categories and conditions.
The notable observation suggests that all three open-vocabulary foundation model-based object detectors, when subjected to degradation of image quality and distribution shift, exhibit significant deviations in performance. This indicates an inherent relationship between OV object detectors and the quality of data. Therefore, this work draws attention to the research community and motivates further research towards improving robustness (refer Figures \ref{fig:meta} and \ref{fig:coco_dc_fig}).

\section{Related Work}
AlexNet \cite{krizhevsky2012imagenet} revolutionized computer vision by applying deep learning, which was further advanced by Fast R-CNN \cite{girshick2015fast} and Faster R-CNN \cite{ren2015faster}, enhancing proposal classification and generation. RetinaNet's \cite{lin2017focal} Focal Loss addressed class imbalance, while YOLO \cite{redmon2016you} achieved real-time detection through a unified regression framework. The Vision Transformer (ViT) \cite{dosovitskiy2020image} innovated by applying transformers to image patches, demonstrating their potential beyond NLP, and the Swin Transformer \cite{liu2021swin} introduced a hierarchical structure for efficient image processing, setting new standards in vision benchmarks.

Carion et al. \cite{carion2020end} developed the DEtection TRansformer (DETR), redefining object detection as a set prediction problem. Utilizing a transformer architecture with an encoder-decoder and a global set-based loss for unique prediction via bipartite matching, DETR streamlines the detection process, removing the necessity for components like non-maximum suppression. However, the detection capability of such models is confined to their trained categories, limiting their effectiveness on unseen objects. To address these constraints, the introduction of foundation models \cite{radford2021learning,li2021grounded} has been proposed.

Foundation AI models like GPT \cite{achiam2023gpt} and BERT \cite{devlin2018bert} in NLP, and CLIP \cite{radford2021learning}, DALL-E \cite{ramesh2021zero}, and SAM \cite{kirillov2023segment} in computer vision, demonstrate the power of large-scale pre-training across data-rich domains. These models have revolutionized tasks like text generation, sentiment analysis, and visual content understanding. Their vast parameter space and training breadth enable extensive adaptability, facilitating the rapid development of specialized AI applications and propelling forward AI research.


Radford et al.'s CLIP \cite{radford2021learning} redefines visual learning with natural language, pre-trained on 400 million web-sourced image-text pairs. This method enables understanding of various visual concepts and supports zero-shot transfer to tasks without specific training. Tested on over 30 benchmarks, including OCR, action recognition, and geo-localization; CLIP's adaptability significantly expands visual model applications.

The Grounded Language-Image Pre-training (GLIP) \cite{li2021grounded} advances visual learning by fusing language supervision, setting new zero-shot and few-shot learning standards. It leverages 27 million annotated and web-sourced data points to enhance object recognition. Open-vocabulary detection, by leveraging NLP, allows models to recognize new objects from descriptions, improving adaptability and robustness in changing environments.

To the best of our knowledge, Zareian et al. \cite{zareian2021open} first open-vocabulary object detection using image-caption pairs for detecting unannotated objects, outperforming zero-shot and weakly supervised approaches by leveraging visual-semantic spaces from captions. Chen et al. \cite{chen2022open} developed MEDet, enhancing Open-Vocabulary Object Detection by aligning vision-language at the proposal level and balancing predictions between known and novel categories, showing top results on MS COCO and LVIS. Bravo et al. \cite{bravo2022localized} proposed a method for open-vocabulary detection that uses localized vision-language matching to improve alignment of visual and linguistic representations, aiming to expand detection vocabularies. Zhao et al. \cite{zhao2022exploiting} demonstrated how combining Vision and Language models with unlabeled data can improve object detection, highlighting the synergy between pre-trained models and large, unlabeled datasets. Zang et al. \cite{zang2022open} introduced OV-DETR, extending DETR for open-vocabulary detection using conditional matching with CLIP-generated embeddings, achieving significant advances on LVIS and COCO. RegionCLIP \cite{zhong2022regionclip} tackled domain shift in open-vocabulary detection by aligning regional visual representations with text, surpassing existing methods on COCO and LVIS for novel and zero-shot categories. Du et al. \cite{du2022learning} presented DetPro, a method for learning continuous prompt representations, enhancing the detection of novel classes through innovative proposal handling and context grading on the LVIS dataset. Feng et al. \cite{feng2022promptdet} created PromptDet, combining CNNs and transformers for object detection, leveraging spatial and global context for improved accuracy and robustness in detection tasks.

Kaul et al. \cite{kaul2023multi} present a multi-modal OVOD method, surpassing the LVIS benchmark by fusing text and visual classifiers with large language models. Cho et al. \cite{cho2023open} innovate in novel object detection using PCL for captions, enhancing LVIS performance via image captioning model distillation. Arandjelovic et al. \cite{arandjelovic2023three} show that combining semantic segmentation with detection improves accuracy in complex scenes. Minderer et al. \cite{minderer2024scaling} explore the benefits of self-training on detection models with large image-text datasets. Shi et al. \cite{shi2023open} develop an OV detection framework based on scene graphs, validated by comprehensive tests. Zhao et al. \cite{zhao2023improving} introduce SAS-Det, a method tackling noisy pseudo labels for improved detection, achieving high COCO and LVIS scores. Kim et al. \cite{kim2023contrastive} propose CFM-ViT, a contrastive learning approach for OV detection that excels on LVIS. Kuo et al. \cite{kuo2022f} demonstrate efficient OV detection with F-VLM by training only the detector head. Kim et al. \cite{kim2023region} present RO-ViT, a pretraining strategy enhancing OV detection alignment, leading to top LVIS and COCO results. Finally, Wang et al. \cite{wang2023object} reveal OADP, an approach for transferring knowledge to OV detectors, outdoing current methods on MS-COCO, and Yao et al. \cite{yao2023detclipv2} introduce DetCLIPv2, a scalable OVD training framework that sets a new zero-shot AP record on LVIS.

Robustness is crucial for assessing the resilience and reliability of machine learning models, emphasizing their ability to perform consistently under variable and unexpected conditions, such as noisy data, distribution shifts, and adversarial attacks. It shows the importance of a model's capacity to generalize beyond its training, ensuring dependable predictions against non-standard inputs. This necessitates a delicate balance between accuracy and generalization, moving away from solely focusing on dataset-specific performance to prevent overfitting. Addressing robustness requires strategies like data augmentation, regularization, and ensemble learning to combat adversarial threats and variability, enhancing model security, privacy, and trustworthiness \cite{hendrycks2021unsolved}. These measures are essential for the practical application of AI in diverse, changing real-world scenarios, aiming to create accurate, secure, and adaptable systems. 
To the best of our knowledge, Hendrycks et al.\cite{hendrycks2019robustness} proposed one of the earliest datasets Imagenet-C to benchmark robustness. Over the years, different versions of datasets have been derived from the original Imagenet dataset to study the topic of robustness for image classification models. Some of the notable datasets are Imagenet-A~\cite{hendrycks2021natural}, Imagenet-R~\cite{hendrycks2021many}, Imagenet-CD~\cite{de2021impact}, Imagenet-E~\cite{li2023imagenet}, Imagenet-X~\cite{idrissi2022imagenet} and Imagenet-Sketch~\cite{wang2019learning}. However, very little work is available in the literature to study the impact of distribution shifts on object detection models.

\section{Out-of-Distribution benchmarks}
In this section, We briefly discuss two OOD benchmarks for the robustness evaluation, namely COCO-O \cite{mao2023coco}, COCO-DC \cite{malik2024objectcompose} and COCO-C \cite{michaelis2019benchmarking}. 
\subsection{COCO-O} Recent robustness benchmark COCO-O~\cite{mao2023coco} dataset dedicated to pose the challenge of object detection under natural distribution shifts, serving as a comprehensive benchmark for assessing detector robustness beyond the typical constraints of existing datasets. 
COCO-O encompasses a range of challenges inherent to object detection, including occlusion, blurring, variations in pose, deformation, illumination differences, and the detection of small-sized objects. COCO-O comprises 6,782 images collected online across six distinct subsets: weather, painting, handmake, cartoon, tattoo, and sketch, arranged in descending order based on the level of detail present within the objects they contain. This organization reflects varying degrees of abstraction across the objects within each domain.

\subsection{COCO-DC} The COCO-DC \cite{malik2024objectcompose} object detection robustness benchmark is recently curated from the COCO 2017 validation set, comprising 1,127 images that distinctly separate foreground objects from their backgrounds. This dataset is designed to evaluate the robustness of object detection and classification models under various background conditions. The COCO-DC benchmark features four subsets: Adversarial, BLIP-2 Caption, Color, and Texture. The Adversarial subset includes images with adversarial background changes crafted to challenge the models' robustness. The BLIP-2 Caption subset utilizes the BLIP-2 model to generate captions for the images, providing a different context for evaluation. The Color subset features images with altered background colors to assess the models' performance under color variations. The Texture subset consists of images with modified background textures to test the models' resilience to texture changes. This benchmark allows for a comprehensive analysis of model performance across diverse and challenging scenarios, highlighting the strengths and weaknesses of contemporary vision-based models in handling object-to-background context variations.

\subsection{COCO-C}  COCO-C \cite{michaelis2019benchmarking} dataset introduces 15 types of image corruptions, each with five levels of severity, covering a broad range of corruption types sorted into four groups: noise, blur, digital, and weather. This comprehensive approach enables a nuanced assessment of model robustness across different distortion types and severity levels, which are not part of the original training regime. The datasets are not intended for training data augmentation but rather to measure a model's robustness against unseen corruptions, thus helping to identify areas for improvement in object detection models.

\section{Open-Vocabulary Object detectors Models}
In this section, we describe the three open-vocabulary foundation models: OWL-ViT, YOLO World, and Grounding DINO. 
\subsection{OWL-ViT} OWL-ViT \cite{minderer2022simple} method introduces an efficient and effective solution for open-vocabulary object detection by leveraging the Vision Transformer (ViT) architecture with minimal modifications and a comprehensive strategy for transferring image-text pre-training to the task of object detection. At its core, OWL-ViT employs a standard ViT for image encoding, which, during the transfer phase to detection, is slightly altered by removing the final token pooling layer and adding lightweight object classification and box prediction heads directly to the output tokens. This design choice allows for the direct prediction of object instances without the need for additional complex mechanisms. The methodology progresses through integrating text embeddings derived from a pretrained language model, facilitating open-vocabulary classification capabilities. This approach strengthens the model to identify a diverse spectrum of object categories beyond those encountered during its initial training phase. Through the end-to-end fine-tuning of both visual and linguistic components on conventional object detection datasets, OWL-ViT showcases exceptional efficacy across various evaluation benchmarks. This performance is attributed to the strategic utilization of extensive image-text corpora for pre-training, succeeded by meticulous fine-tuning processes. Consequently, OWL-ViT establishes new benchmarks in the domains of zero-shot, text-conditioned, and one-shot, image-conditioned object detection tasks.
\subsection{YOLO-World} YOLO-World \cite{cheng2024yolow} introduces a novel open-vocabulary object detection framework that significantly enhances the conventional YOLO detection model with the capacity for open-vocabulary detection, achieving real-time efficiency and high accuracy across diverse benchmarks. YOLO-World incorporates the Re-parameterizable Vision-Language Path Aggregation Network (RepVL-PAN) and a distinct region-text contrastive loss. These components work in tandem to ensure a robust visual-semantic alignment between image features and textual embeddings. This strategic integration strengthens the model's ability to adeptly navigate the complex interplay between visual and linguistic information, significantly enhancing its open-vocabulary detection capabilities while maintaining real-time processing speeds. This approach leverages large-scale datasets for pre-training, effectively combining detection, grounding, and image-text data into a unified learning framework. As a result, YOLO-World not only extends the capabilities of the YOLO architecture to recognize a broader array of object categories in a zero-shot manner but also does so with remarkable inference speed and deployment efficiency. 
\subsection{Grounding DINO} Grounding DINO \cite{liu2023grounding} a novel approach to open-set object detection by integrating the strengths of the Transformer-based detector DINO with grounded pre-training techniques. This method allows for detecting arbitrary objects based on human input, such as category names or referring expressions, by effectively fusing language and vision modalities. Grounding DINO partitions a closed-set detector into three phases—feature enhancement, language-guided query selection, and cross-modality decoder—to achieve a tightly integrated fusion of language and vision. Unlike prior works that evaluate open-set object detection primarily on novel categories, Grounding DINO also extends evaluations to referring expression comprehension (REC), enabling the model to understand objects specified with attributes. 
\section{Experiments and Results}
\label{sec:results}
The list of the object categories present in the input image is used as a text query for zero-shot evaluation of the open vocabulary object detection models (refer to Figure \ref{fig:cartoon}). This work evaluates OWL-ViT, YOLO-World, and Grounding DINO models on all six OOD subsets of COCO-O, 5 subsets of COCO-DC \cite{malik2024objectcompose}, and on fifteen corruption subsets (at severity level 1) of COCO-C benchmark \cite{hendrycks2021many}.

\subsection{Evaluation Method}
\begin{figure}[h]
\vspace{-4mm}
  \centering
  \includegraphics[width=0.8\columnwidth]{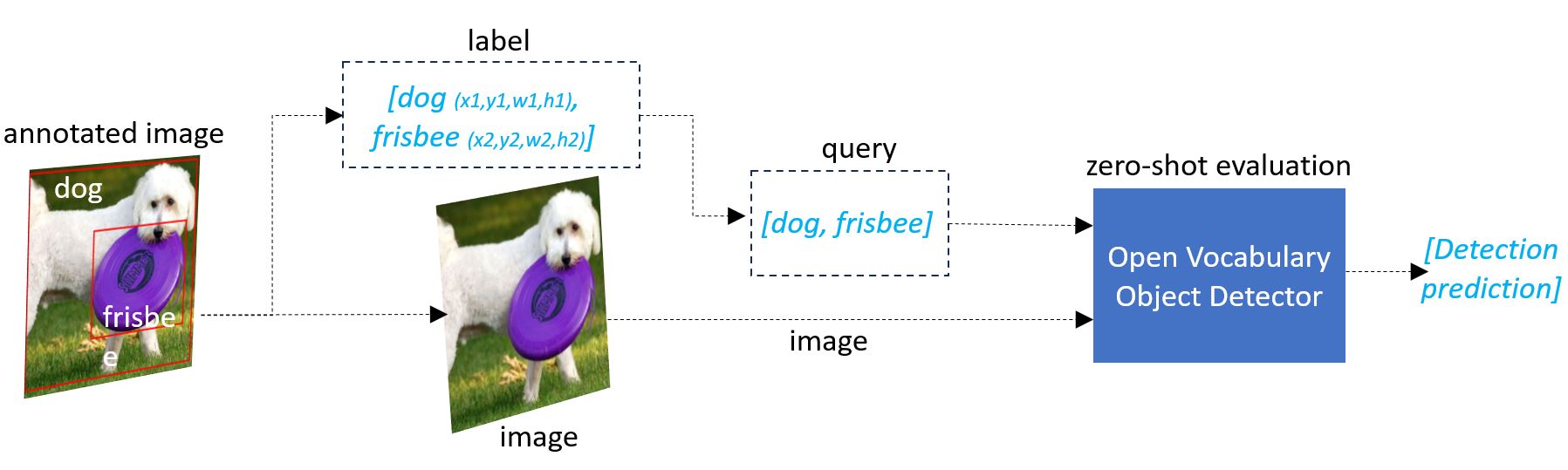}
  \vspace{-4mm}
  \caption{Zero-shot evaluation process of open vocabulary object detector models.}
  \label{fig:ovod}
  \vspace{-4mm}
\end{figure}

This work evaluates the robustness of OWL-ViT, YOLO World, and Grounding DINO models for their zero-shot capability based on their performance on OOD benchmarks COCO-O, COCO-DC, and COCO-C, as shown in Figure \ref{fig:ovod}.
\subsection{Metrics}
AP (Average Precision) measures the precision of a model at different confidence thresholds, summarizing its detection accuracy for a specific class. It is the area under the precision-recall curve.

mAP@IoU=0.5 (mean Average Precision at Intersection over Union of 0.5) averages the AP values for all classes at an IoU threshold of 0.5, meaning detections are correct if the predicted and ground truth boxes overlap by at least 50\%. Box mAP@IoU=0.5:0.95 evaluates the model across IoU thresholds from 0.5 to 0.95, in steps of 0.05. It calculates the AP at each threshold and averages these values, providing a rigorous assessment of detection precision at varying overlap levels.

The effective robustness metric $ER(f)$ for a detector $f$ is computed as given in Eq. \ref{eq:ER}. $mAP_{ood}$ is the $mAP$ on COCO-O dataset, and $mAP_{id}$ is the $mAP$ on original COCO dataset. $mAP_{id}$ is multiplied by a factor $\beta$ in the equation. The value of $\beta$ is fixed to $0.45$, adopted from \cite{mao2023coco}. 
\begin{equation}
    ER(f)=mAP_{ood}(f)-\beta \times mAP_{id}(f) \label{eq:ER}
\end{equation}

\subsection{Discussions}
Detailed results on six subsets of COCO-O benchmarks are described in Table \ref{tab:compare} and \ref{tab:compare-05}, results on subsets of COCO-DC benchmark is in Table \ref{tab:coco_dc}, and for 15 corruptions subsets of COCO-C benchmarks, results are in Table  \ref{tab:cococ_weather}, \ref{tab:cococ_blur}, \ref{tab:cococ_digital}, and \ref{tab:cococ_noise}.

\begin{table}[ht]
\centering
\caption{Comparison of Box mAP (@IoU=0.50:0.95) for different detectors in original COCO and COCO-O datasets. $^*$referred from Grounding DINO \cite{liu2023grounding}.}
\tiny
\label{tab:compare}
\begin{tabular}{ccccccccccc}
\hline
\multicolumn{11}{c}{\textbf{Box mAP@IoU=0.50:0.95}} \\ \hline
\textbf{} & \textbf{Venue} & \textbf{\begin{tabular}[c]{@{}c@{}}COCO\\ mAP\end{tabular}} & \multicolumn{7}{c}{\textbf{\begin{tabular}[c]{@{}c@{}}COCO-O\\ mAP\end{tabular}}} & \multicolumn{1}{c}{\textbf{\begin{tabular}[c]{@{}c@{}}\end{tabular}}} \\ \hline
\textbf{} & \textbf{} & \textbf{} & \textbf{Sketch} & \textbf{Weather} & \textbf{Cartoon} & \textbf{Painting} & \textbf{Tattoo} & \textbf{Handmake} & \multicolumn{1}{c}{\textbf{Average}} \\ \hline
OWL-ViT & ECCV'22 & 26.40 & 14.60 & 19.50 & 17.20 & 24.50 & 7.70 & 12.30 & 15.97 \\
YOLO World & CVPR'24 & 39.30 & 15.00 & 37.90 & 18.40 & 36.00 & 10.10 & 23.10 & 23.42 \\
Grounding DINO & ECCV'24 & 48.40$^*$ & 44.90 & 33.70 & 47.50 & 42.30 & 41.10 & 39.60 & 41.52 \\ \hline
\end{tabular}
\vspace{-3mm}
\end{table}

\begin{table*}[ht]
\centering
\caption{Comparison of mAP (@IoU=0.50) for different detectors in original COCO and COCO-O datasets.}
\tiny
\label{tab:compare-05}
\begin{tabular}{cccccccccc}
\hline
\multicolumn{10}{c}{\textbf{mAP@IoU=0.50}} \\ \hline
\textbf{} & \textbf{Venue} & \textbf{\begin{tabular}[c]{@{}c@{}}COCO\\ mAP\end{tabular}} & \multicolumn{7}{c}{\textbf{\begin{tabular}[c]{@{}c@{}}COCO-O\\ mAP\end{tabular}}} \\ \hline
\textbf{} & \textbf{} & \textbf{} & \textbf{Sketch} & \textbf{Weather} & \textbf{Cartoon} & \textbf{Painting} & \textbf{Tattoo} & \textbf{Handmake} & \textbf{Average} \\ \hline
OWL-ViT & ECCV'22 & 42.90 & 18.90 & 31.00 & 23.50 & 33.30 & 9.80 & 16.10 & 22.10 \\
YOLO World & CVPR'24 & 51.20 & 17.30 & 46.00 & 22.50 & 43.80 & 12.40 & 27.40 & 28.23 \\
Grounding DINO & ECCV'24 & - & 53.00 & 40.90 & 55.90 & 49.50 & 50.90 & 45.20 & 49.23 \\ \hline
\end{tabular}
\vspace{-2mm}
\end{table*}

\begin{figure}[h]
\vspace{-4mm}
  \centering
  \includegraphics[width=0.5\columnwidth]{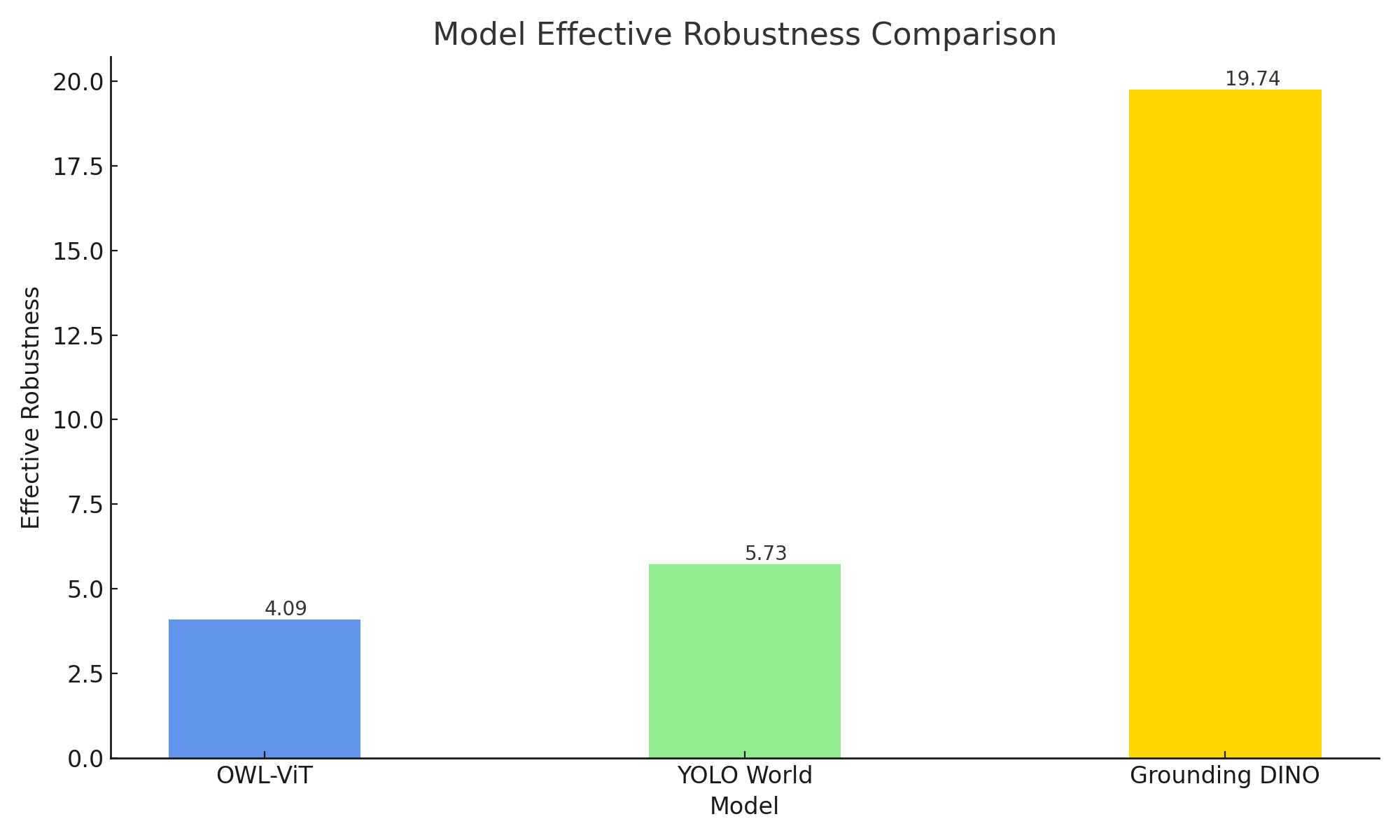}
  \vspace{-4mm}
  \caption{Comparisons of effective robustness for detectors based on their performance on original COCO and COCO-O datasets.}
  \label{fig:er}
\end{figure}

\textbf{COCO-O}: Following the results in Table \ref{tab:compare} and \ref{tab:compare-05}, OWL-ViT on the original COCO dataset stands at 26.40 and 42.90 for IoU=0.50:0.95 and IoU=0.50, respectively. When subjected to the COCO-O dataset, the average performance drops to 15.97 and 22.10 across the same IoU thresholds, indicating a significant decrease in robustness under out-of-distribution conditions. 
\begin{figure}[ht]
  \centering
  \includegraphics[width=0.7\columnwidth]{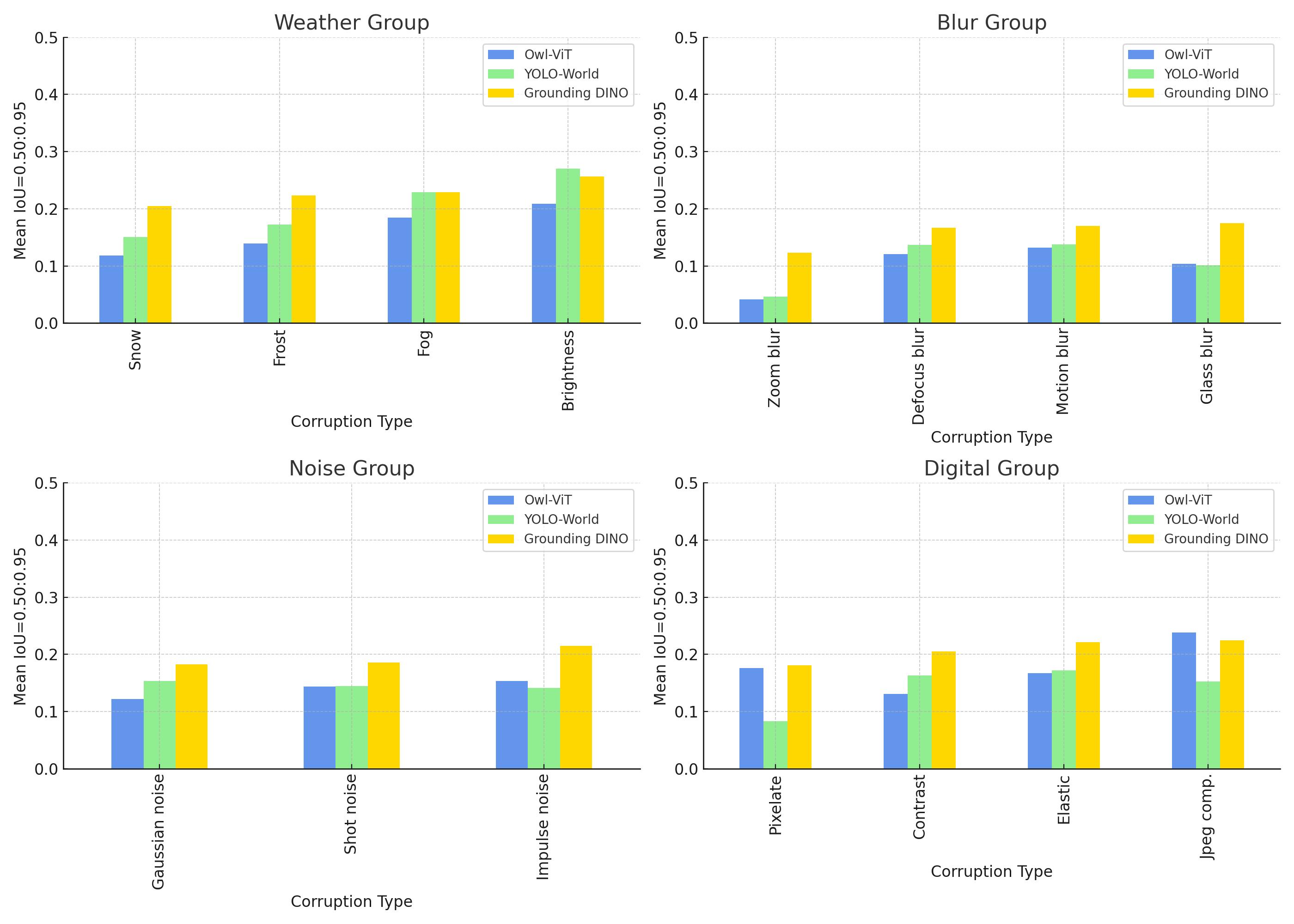}
  \caption{Comparisons of performance of the models (Owl-ViT, YOLO-World, and Grounding DINO) on four corruptions group of COCO-C.}
  \label{fig:coco_c_groupwise}
\end{figure}
Similarly, YOLO World, which achieves a higher baseline mAP of 39.30 and 51.20 on the COCO dataset for IoU=0.50:0.95 and IoU=0.50, respectively, shows a reduced average mAP of 23.42 and 28.23 on COCO-O. This suggests a resilience to out-of-distribution data compared to OWL-ViT, though the performance still notably decreases. Grounding DINO demonstrates the most remarkable performance, with a mAP of 48.40 on COCO for IoU=0.50:0.95 and a consistent lack of data for IoU=0.50. On COCO-O, it achieves an average mAP of 41.52 and 49.23 across the respective IoU thresholds. Grounding DINO exhibits the least performance drop among the three, indicating robustness to out-of-distribution scenarios, as indicated the trend of effective robustness of models in Fig. \ref{fig:er}.

\textbf{COCO-DC}: The performance of three open-vocabulary object detectors—OWL-ViT, YOLO World, and Grounding DINO—on the COCO-DC dataset subsets is summarized in Table~\ref{tab:coco_dc}. All three models exhibit high performance on the Original subset, with YOLO World achieving the highest Box mAP (49.8), followed by OWL-ViT (42.5) and Grounding DINO (40.4). However, the Adversarial subset causes a significant drop for all models. Grounding DINO demonstrates the highest robustness (34.7), while YOLO World and OWL-ViT drop to 24.6 and 22.8, respectively. On the BLIP-2 Caption subset, Grounding DINO (42.6) and YOLO World (39.5) perform well, but OWL-ViT lags (34.2).

\begin{table}[ht]
\centering
\caption{Comparison of Box mAP (@IoU=0.50:0.95) for different detectors in the COCO-DC dataset subsets.}
\scriptsize
\label{tab:coco_dc}
\begin{tabular}{cccccc}
\hline
\multicolumn{6}{c}{\textbf{Box mAP@IoU=0.50:0.95}} \\ \hline
\textbf{} & \textbf{Original} & \textbf{Adversarial} & \textbf{BLIP} & \textbf{Color} & \textbf{Texture} \\ \hline
OWL-ViT & 42.5 & 22.8 & 34.2 & 33.4 & 30.6 \\
YOLO World & 49.8 & 24.6 & 39.5 & 29.3 & 28.4 \\
Grounding DINO & 40.4 & 34.7 & 42.6 & 44.4 & 43.9 \\ \hline
\end{tabular}
\vspace{-3mm}
\end{table}

The Color and Texture subsets challenge all models, with pronounced performance drops. Grounding DINO maintains the highest scores (44.4 and 43.9), showing superior adaptability. YOLO World scores 29.3 for Color and 28.4 for Texture, while OWL-ViT scores 33.4 and 30.6, respectively. This suggests Grounding DINO's training strategy allows better generalization across visual distortions. Overall, while Grounding DINO exhibits the highest resilience, all models show vulnerabilities under distribution shifts, emphasizing the need for diverse and challenging training scenarios to enhance robustness in real-world applications.

\begin{table*}[!ht]
\centering
\caption{Robustness evaluation on Weather group: Snow, Frost, Fog, and Brightness corruption subsets in the COCO-C dataset.}
\tiny
\label{tab:cococ_weather}
\begin{tabular}{cccccccc}
\hline
Corruption & Severity & \multicolumn{2}{c}{Owl-ViT} & \multicolumn{2}{c}{YOLO-World} & \multicolumn{2}{c}{Grounding DINO} \\
\cline{3-8} 
 & & IoU=0.50:0.95 & IoU=0.50 & IoU=0.50:0.95 & IoU=0.50 & IoU=0.50:0.95 & IoU=0.50 \\
\hline
Snow & 1 & 18.0 & 29.9 & 25.2 & 32.7 & 24.7 & 34.4 \\
Snow & 2 & 12.8 & 20.6 & 14.4 & 19.3 & 21.2 & 30.2 \\
Snow & 3 & 11.9 & 19.8 & 14.0 & 18.8 & 20.4 & 28.8 \\
Snow & 4 & 8.3 & 13.7 & 10.9 & 14.8 & 18.6 & 27.3 \\
Snow & 5 & 8.2 & 13.4 & 10.8 & 14.6 & 17.3 & 25.5 \\
\hline
Frost & 1 & 19.8 & 32.4 & 28.9 & 37.5 & 24.6 & 33.9 \\
Frost & 2 & 15.5 & 25.2 & 17.8 & 23.6 & 22.6 & 31.9 \\
Frost & 3 & 12.2 & 19.9 & 14.6 & 19.4 & 20.5 & 29.2 \\
Frost & 4 & 12.1 & 19.4 & 14.0 & 18.5 & 20.3 & 28.8 \\
Frost & 5 & 10.1 & 16.4 & 11.7 & 15.5 & 18.9 & 27.5 \\
\hline
Fog & 1 & 20.9 & 33.7 & 29.8 & 38.4 & 23.9 & 32.7 \\
Fog & 2 & 19.6 & 31.6 & 22.6 & 29.8 & 23.4 & 32.0 \\
Fog & 3 & 18.6 & 29.8 & 21.5 & 28.3 & 23.0 & 31.4 \\
Fog & 4 & 17.6 & 28.2 & 21.3 & 28.1 & 22.7 & 31.1 \\
Fog & 5 & 15.4 & 24.5 & 19.6 & 25.8 & 21.9 & 30.3 \\
\hline
Brightness & 1 & 23.3 & 37.8 & 33.5 & 43.5 & 26.1 & 35.5 \\
Brightness & 2 & 22.4 & 36.2 & 27.0 & 35.6 & 26.1 & 35.6 \\
Brightness & 3 & 21.2 & 34.2 & 26.2 & 34.6 & 25.9 & 35.5 \\
Brightness & 4 & 19.6 & 31.4 & 25.0 & 33.1 & 25.5 & 35.0 \\
Brightness & 5 & 17.9 & 28.5 & 23.4 & 31.0 & 24.9 & 34.5 \\
\hline
Original COCO & - & 26.40 & 42.90 & 39.30 & 51.20 & 48.40 & - \\
\hline
\end{tabular}
\vspace{-2mm}
\end{table*}

\begin{table*}[ht]
\centering
\caption{Robustness evaluation on Blur group: Zoom blur, Defocus blur, Motion blur, and Glass blur corruption subsets in the COCO-C dataset.}
\tiny
\label{tab:cococ_blur}
\begin{tabular}{cccccccc}
\hline
Corruption & Severity & \multicolumn{2}{c}{Owl-ViT} & \multicolumn{2}{c}{YOLO-World} & \multicolumn{2}{c}{Grounding DINO} \\
\cline{3-8} 
 & & IoU=0.50:0.95 & IoU=0.50 & IoU=0.50:0.95 & IoU=0.50 & IoU=0.50:0.95 & IoU=0.50 \\
\hline
Zoom blur & 1 & 8.3 & 15.8 & 12.6 & 20.8 & 13.3 & 22.6 \\
Zoom blur & 2 & 5.1 & 10.5 & 4.7 & 8.8 & 10.0 & 18.6 \\
Zoom blur & 3 & 3.4 & 7.4 & 3.0 & 6.0 & 8.1 & 16.1 \\
Zoom blur & 4 & 2.2 & 5.1 & 1.8 & 3.9 & 6.4 & 13.5 \\
Zoom blur & 5 & 1.6 & 3.9 & 1.2 & 2.7 & 5.4 & 11.9 \\
\hline
Defocus blur & 1 & 17.9 & 28.1 & 28.0 & 36.3 & 20.7 & 28.2 \\
Defocus blur & 2 & 15.5 & 24.1 & 19.4 & 26.0 & 18.8 & 26.2 \\
Defocus blur & 3 & 11.9 & 18.4 & 14.0 & 19.2 & 16.7 & 23.9 \\
Defocus blur & 4 & 9.1 & 14.1 & 9.1 & 12.7 & 14.2 & 20.5 \\
Defocus blur & 5 & 6.9 & 10.8 & 5.1 & 7.2 & 12.8 & 18.7 \\
\hline
Motion blur & 1 & 19.4 & 31.8 & 27.9 & 37.3 & 21.8 & 30.1 \\
Motion blur & 2 & 16.2 & 26.7 & 17.3 & 24.5 & 19.5 & 26.2 \\
Motion blur & 3 & 12.0 & 20.3 & 11.0 & 16.0 & 16.0 & 22.0 \\
Motion blur & 4 & 7.8 & 13.3 & 5.2 & 7.9 & 12.5 & 19.1 \\
Motion blur & 5 & 5.5 & 9.2 & 2.8 & 4.4 & 10.6 & 18.2 \\
\hline
Glass blur & 1 & 17.3 & 26.9 & 25.7 & 33.7 & 21.1 & 28.5 \\
Glass blur & 2 & 14.2 & 22.0 & 15.0 & 20.1 & 15.4 & 19.8 \\
Glass blur & 3 & 8.3 & 12.9 & 4.5 & 6.3 & 12.6 & 17.6 \\
Glass blur & 4 & 6.8 & 10.5 & 3.1 & 4.3 & 11.2 & 15.7 \\
Glass blur & 5 & 5.1 & 7.9 & 1.9 & 2.6 & 9.3 & 13.5 \\
\hline
Original COCO & - & 26.40 & 42.90 & 39.30 & 51.20 & 48.40 & - \\
\hline
\end{tabular}
\vspace{-3mm}
\end{table*}

\begin{table*}[!ht]
\centering
\caption{Robustness evaluation on Noise group: Gaussian noise, Shot noise, and Impulse noise corruption subsets in the COCO-C dataset.}
\tiny
\label{tab:cococ_noise}
\begin{tabular}{cccccccc}
\hline
Corruption & Severity & \multicolumn{2}{c}{Owl-ViT} & \multicolumn{2}{c}{YOLO-World} & \multicolumn{2}{c}{Grounding DINO} \\
\cline{3-8} 
 & & IoU=0.50:0.95 & IoU=0.50 & IoU=0.50:0.95 & IoU=0.50 & IoU=0.50:0.95 & IoU=0.50 \\
\hline
Gaussian noise & 1 & 22.4 & 36.8 & 28.2 & 36.8 & 21.8 & 29.6 \\
Gaussian noise & 2 & 19.3 & 32.0 & 23.4 & 31.0 & 19.3 & 26.2 \\
Gaussian noise & 3 & 14.9 & 24.7 & 16.5 & 22.0 & 19.4 & 27.3 \\
Gaussian noise & 4 & 9.9 & 16.5 & 8.5 & 11.4 & 17.7 & 25.9 \\
Gaussian noise & 5 & 4.6 & 7.5 & 2.1 & 2.8 & 12.7 & 19.1 \\
\hline
Shot noise & 1 & 22.7 & 37.4 & 28.4 & 37.1 & 22.2 & 29.9 \\
Shot noise & 2 & 19.3 & 31.9 & 18.2 & 24.5 & 20.2 & 27.5 \\
Shot noise & 3 & 14.8 & 24.4 & 12.5 & 17.0 & 18.5 & 24.7 \\
Shot noise & 4 & 9.4 & 15.4 & 4.7 & 6.5 & 15.2 & 21.0 \\
Shot noise & 5 & 5.5 & 8.9 & 1.7 & 2.4 & 12.1 & 18.0 \\
\hline
Impulse noise & 1 & 19.6 & 32.2 & 25.8 & 33.8 & 22.7 & 30.9 \\
Impulse noise & 2 & 17.1 & 28.0 & 15.6 & 20.9 & 21.4 & 29.2 \\
Impulse noise & 3 & 14.7 & 24.3 & 11.7 & 15.9 & 20.4 & 28.6 \\
Impulse noise & 4 & 9.9 & 16.5 & 4.5 & 6.1 & 18.4 & 26.7 \\
Impulse noise & 5 & 5.1 & 8.5 & 0.9 & 1.3 & 13.4 & 20.1 \\
\hline
Original COCO & - & 26.40 & 42.90 & 39.30 & 51.20 & 48.40 & - \\
\hline
\end{tabular}
\vspace{-3mm}
\end{table*}
\begin{table*}[ht]
\centering
\caption{Robustness evaluation on Digital group: Pixelate, Contrast, Elastic Transform, and Jpeg compression corruption subsets in the COCO-C dataset.}
\tiny
\label{tab:cococ_digital}
\begin{tabular}{cccccccc}
\hline
Corruption & Severity & \multicolumn{2}{c}{Owl-ViT} & \multicolumn{2}{c}{YOLO-World} & \multicolumn{2}{c}{Grounding DINO} \\
\cline{3-8} 
 & & IoU=0.50:0.95 & IoU=0.50 & IoU=0.50:0.95 & IoU=0.50 & IoU=0.50:0.95 & IoU=0.50 \\
\hline
Pixelate & 1 & 24.8 & 40.0 & 23.4 & 30.2 & 23.3 & 31.1 \\
Pixelate & 2 & 24.3 & 39.1 & 13.5 & 17.8 & 21.5 & 28.4 \\
Pixelate & 3 & 17.7 & 28.0 & 4.9 & 6.5 & 17.5 & 23.3 \\
Pixelate & 4 & 12.8 & 19.9 & 1.9 & 2.7 & 13.8 & 18.4 \\
Pixelate & 5 & 8.5 & 12.6 & 0.6 & 0.7 & 9.9 & 13.5 \\
\hline
Contrast & 1 & 21.1 & 33.9 & 29.7 & 38.3 & 23.7 & 32.3 \\
Contrast & 2 & 19.5 & 30.9 & 21.8 & 28.7 & 23.0 & 31.3 \\
Contrast & 3 & 16.3 & 25.6 & 18.4 & 24.1 & 22.0 & 30.0 \\
Contrast & 4 & 9.2 & 14.1 & 9.4 & 12.5 & 16.0 & 24.0 \\
Contrast & 5 & 2.3 & 3.5 & 1.6 & 2.0 & 11.0 & 16.1 \\
\hline
Elastic & 1 & 21.6 & 35.3 & 27.9 & 36.7 & 24.0 & 33.5 \\
Elastic & 2 & 19.4 & 32.3 & 19.1 & 26.2 & 22.8 & 32.3 \\
Elastic & 3 & 16.8 & 28.2 & 14.8 & 20.7 & 20.5 & 30.1 \\
Elastic & 4 & 14.4 & 24.4 & 11.7 & 16.5 & 19.3 & 28.2 \\
Elastic & 5 & 11.6 & 20.0 & 8.2 & 11.8 & 17.2 & 25.5 \\
\hline
Jpeg comp. & 1 & 27.3 & 44.6 & 24.9 & 32.5 & 24.3 & 33.6 \\
Jpeg comp. & 2 & 26.9 & 44.3 & 15.2 & 20.7 & 22.4 & 30.6 \\
Jpeg comp. & 3 & 26.5 & 44.1 & 12.4 & 16.7 & 19.0 & 26.2 \\
Jpeg comp. & 4 & 21.4 & 36.5 & 6.8 & 9.3 & 17.3 & 24.3 \\
Jpeg comp. & 5 & 13.0 & 22.4 & 2.6 & 3.5 & 14.3 & 20.3 \\
\hline
Original COCO & - & 26.40 & 42.90 & 39.30 & 51.20 & 48.40 & - \\
\hline
\end{tabular}
\vspace{-2mm}
\end{table*}
\begin{figure}[!ht]
  \centering
  \includegraphics[width=0.7\columnwidth]{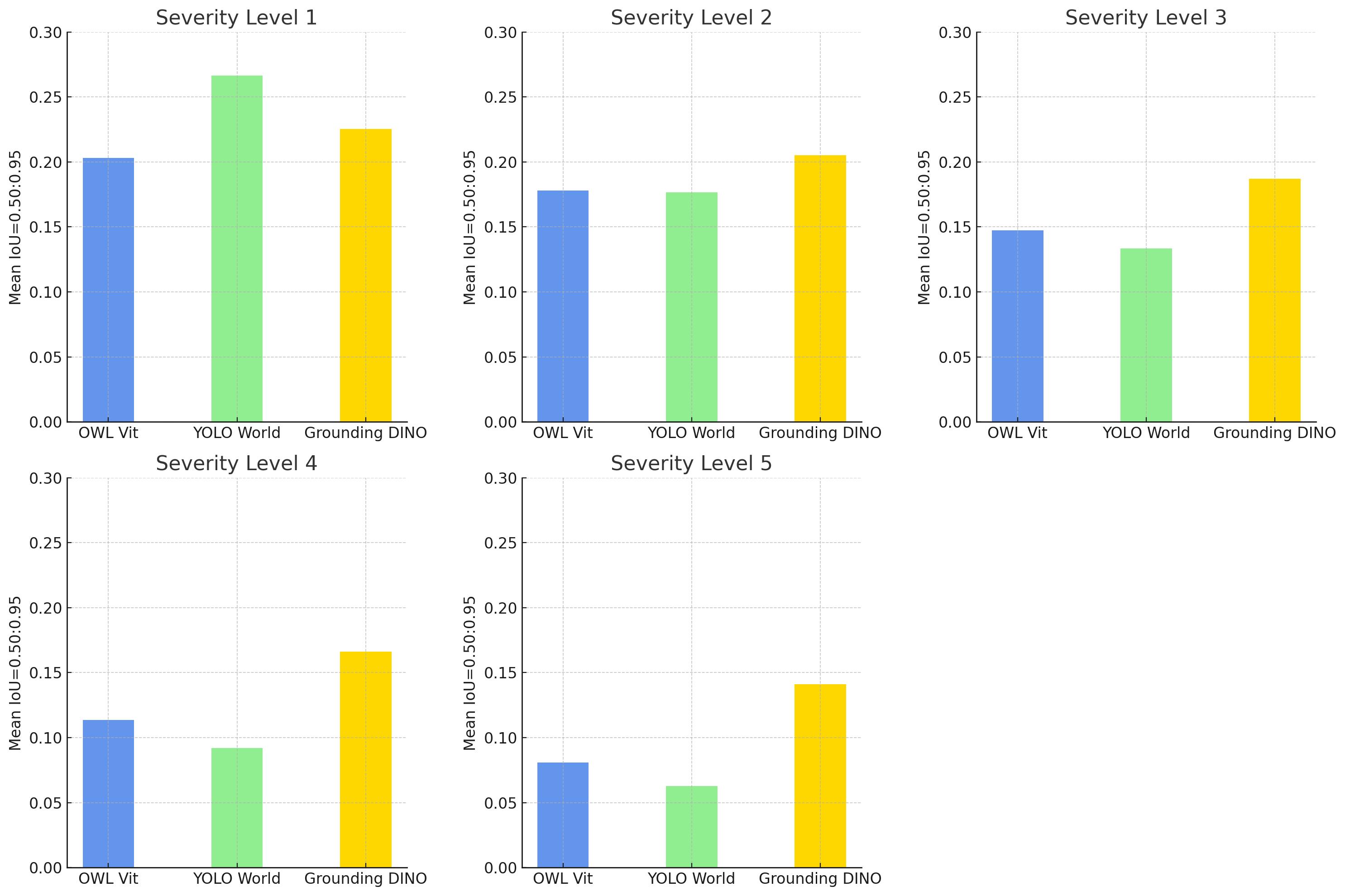}
  \caption{Comparisons of performance of the models (Owl-ViT, YOLO-World, and Grounding DINO) on across severity level of underlying corruptions in COCO-C.}
  \label{fig:coco_c_levelwise}
  \vspace{-2mm}
\end{figure}

\textbf{COCO-C}: In the robustness evaluation conducted on the COCO-C dataset, three models---Owl-ViT, YOLO-World, and Grounding DINO were assessed across fifteen different corruption types grouped into Weather, Blur, Noise, and Digital categories (detailed results in Table \ref{tab:cococ_weather}, \ref{tab:cococ_blur}, \ref{tab:cococ_noise}, and \ref{tab:cococ_digital}).

The results reveal a trend of decreasing performance with increasing severity of corruption across all models and corruption types, highlighted by notable performance drops such as Owl-ViT’s decrease from 40.0 to 12.6 IoU in Pixelate corruption from severity 1 to 5. 

YOLO-World generally displayed superior resilience in lower severity levels, particularly in handling weather-related and blur corruptions, where it outperformed Owl-ViT, managing a 37.5 IoU at severity 1 for Frost compared to Owl-ViT's 32.4. Conversely, Grounding DINO, while slightly underperforming in higher intersection over union (IoU) metrics, showed comparable or better performance in detecting objects at the basic IoU=0.50 level across many corruption scenarios, such as achieving an IoU of 35.5 in high-severity Brightness corruption versus Owl-ViT’s 28.5. 
As trend shown in Figure  \ref{fig:coco_c_groupwise}, across all groups suggests that while Grounding DINO often excels in handling blur-induced distortions, Owl-ViT generally lags behind, especially as the complexity of corruptions increases.  A meta-comparison (Fig. \ref{fig:meta} reveals varying degrees of resilience among the models. Grounding DINO stands out for its robustness, maintaining closer performance levels between the original and out-of-distribution benchmarks. YOLO-World shows moderate resilience, with a noticeable but smaller performance drop compared to OWL-ViT, which experiences the most significant decrease in mAP when transitioning from COCO to OOD benchmarks.

\noindent \textbf{Increased severity levels:} Figure \ref{fig:coco_c_levelwise} illustrates the decline in performance across five severity levels for the Owl-ViT, YOLO-World, and Grounding DINO models. YOLO-World consistently outperforms the other models across all severity levels, maintaining a higher mean IoU, particularly at severity level 1 where it achieves a performance peak notably higher than its counterparts. As severity increases, all models demonstrate a downward trend, with Owl-ViT having a substantial drop, especially notable between severity levels 1 and 4. Grounding DINO, while not leading at lower severities, shows a more gradual decline, suggesting a degree of robustness in more challenging conditions, as its performance at severity level 5 remains competitive with YOLO-World's. The consistency of these results with the earlier detailed performance metrics across various corruption types validates the trend that model robustness significantly diminishes with increased corruption severity. This emphasizes the importance of robustness evaluations across different levels of corruption severity to assess the reliability of models in real-world scenarios. These findings emphasize the need for further research on robust models capable of maintaining performance under varying degrees of corruption. The consistent performance drop across models points to an essential area for future research: enhancing object detection models' adaptability and resilience against varied and newer visual concept.

\section{Conclusion}
To the best of our knowledge, we are making one of the earliest attempts to understand zero-shot evaluation on open-vocabulary foundation models through the perspective of robustness under distribution shifts. 
Through extensive analysis of three recent open-vocabulary foundation object detection models on three public benchmarks, we show that object detection under conditions of out-of-distribution (OOD) shifts poses significant challenges regarding performance deviation, advocating increased focus and investigation by the research community. 
Using vision-language models combined with effective, prompt engineering can be the future direction for developing more robust open-vocabulary object detectors. Enhancing robustness against various distribution shifts increases the trustworthiness of open-vocabulary object detection models, potentially leading to their adoption across diverse applications.

\section{Acknowledgment} The authors thank Sumit Rakesh, Luleå University of Technology, for his support with the Lotty Bruzelius cluster. We also thank the National Supercomputer Centre at Linköping University for the Berzelius supercomputing, supported by the Knut and Alice Wallenberg Foundation.
%
%
%
\newpage
\bibliographystyle{splncs04}
\bibliography{main}

\end{document}